\pdfoutput=1
\documentclass[final]{cvpr}
\usepackage{times}
\usepackage{epsfig}
\usepackage{graphicx}
\usepackage{amsmath}
\usepackage{amssymb}

\usepackage{array,booktabs,calc}
\usepackage{placeins}
\usepackage[inline, shortlabels]{enumitem}
\usepackage{graphicx}
\usepackage{multirow}
\usepackage{wrapfig}
\usepackage{paralist}
\usepackage[dvipsnames]{xcolor}
\usepackage{inconsolata}
\usepackage{gensymb}
\usepackage{caption} 
\usepackage{bbm}  %
\usepackage{xcolor}
\usepackage{floatrow}
\usepackage{nicefrac}
\usepackage{subfig}
\usepackage{stfloats}
\usepackage[pagebackref, pageanchor=true, plainpages=false, pdfpagelabels, bookmarks, bookmarksnumbered]{hyperref}

\setlist{itemjoin ={,\enspace},itemjoin* = { and\enspace}}
\definecolor{citecolor}{HTML}{0071bc}
\hypersetup{
  breaklinks   = true,
  colorlinks   = true, %
  urlcolor     = RubineRed, %
  linkcolor    = RubineRed, %
  citecolor    = citecolor %
}
\renewcommand{\b}[1]{\textbf{#1}}

\begin{document}

\title{Keypoints Tracking via Transformer Networks}

\author{
    Oleksii Nasypanyi$^{1}$ 
    \quad Francois Rameau$^{1}$\\[1.5mm]
    $^1$KAIST \quad 
}

\maketitle

\begin{abstract}
  \textit{
In this thesis, we propose a pioneering work on sparse keypoints tracking across images using transformer networks. While deep learning-based keypoints matching have been widely investigated using graph neural networks - and more recently transformer networks, they remain relatively too slow to operate in real-time and are particularly sensitive to the poor repeatability of the keypoints detectors. In order to address these shortcomings, we propose to study the particular case of real-time and robust keypoints tracking. 
Specifically, we propose a novel architecture which ensures a fast and robust estimation of the keypoints tracking between successive images of a video sequence. Our method takes advantage of a recent breakthrough in computer vision, namely, visual transformer networks. Our method consists of two successive stages, a coarse matching followed by a fine localization of the keypoints’ correspondences prediction. Through various experiments, we demonstrate that our approach achieves competitive results and demonstrates high robustness against adverse conditions, such as illumination change, occlusion and viewpoint differences. 
Code is available at our project page: \url{https://github.com/LexaNagiBator228/Keypoints-Tracking-via-Transformer-Networks/}.
}

\end{abstract}

\section{Introduction}\label{sec:intro}
For decades keypoints matching and tracking have been a cornerstone for a large spectrum of applications such as SLAM[4], visual odometry[6], motion detection[7], time to collision[8], place localization[9], and more. Traditional approaches for keypoints matching [10, 11] and tracking relies on handcrafted features extracted locally in the images. While these approaches have demonstrated their versatility and effectiveness, they remain very sensitive 
to a large number of factors such as illumination changes and viewpoints differences between views. To cope with these limitations, recent deep learning based techniques have emerged[12, 13].

\begin{figure}[tb]
    \centering
    \includegraphics[width=0.95\linewidth]{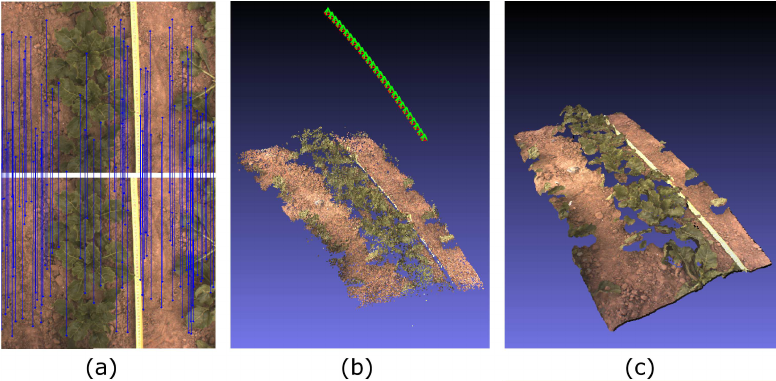}
    \caption[ Illustration of using keypoints matching for 3D reconstruction.]{
        \textbf{Illustration of using keypoints matching for 3D reconstruction [22].}
    }
    \label{fig:teaser}
\end{figure}

The first attempts to replace the traditional visual odometry pipeline by neural network approaches were focusing on the end-to-end learning of the camera motion with limited intermediate supervision [14]. While the implementation of these techniques is straightforward and intuitive, their generalization to arbitrary scenes remains debatable. This problem was partly resolved by self-supervised visual-odometry approaches [15] which can be easily trained on unlabelled video sequences. Despite this advantage, their poor transferability drastically limits the deployment of these end-to-end strategies.  
For the affordmention reason, replacing the visual odometry pipeline in its entirety is questionable. To compute a camera motion a few stages are required: keypoints detection/description/matching and  robust geometric estimation (often achieved by a RANSAC fitting a parametric motion model associated with a non-linear refinement). While the second stage is well mathematically defined and understood, the first step (keypoint extraction and matching) is not fully resolved. In this context, performing the local features extraction and association using deep neural networks is very relevant. Unsurprisingly, many attempts have already been proposed, for instance, a seminal work entitled LIFT[16] learns keypoints detection and description with a large dataset of images acquired under various conditions. More recently, SuperPoints[2], and D2Net[17] have demonstrated better performances by learning keypoints detection and description via synthetically generated data or RGB-D datasets (Megadepth[18], ScanNet[19] etc.). These CNN based approaches undeniably improved the robustness of keypoints association between pairs of images (Show example of results from recent approaches and in the figure caption put the challenge it resolved like viewpoint difference, illumination etc.)
Although detecting keypoints on successive images before matching them is a commonly employed strategy for 3D reconstruction, it is not necessarily the most effective technique when dealing with video for various reasons. First these approaches do not consider the sequential nature of the data. Secondly, the repeatability of the keypoints detection can lead to a large number of wrongly - or missing- matches which can be problematic under complex scenarios.
To cope with these limitations, we propose to explore the particular problem of keypoints tracking using neural networks which has, so far, attracted a relatively narrow attention. Specifically, we propose to resolve this problem using a very recent deep learning architecture called transformer networks[20] which happen to be very well suited for the task at hand. Our approach is hierarchical since a coarse keypoint tracking is accurately refined by a second transformer network. Also we developed our own strategy to deal with outliers, and a special training pipeline for it.
In our experiments we compared the matching accuracy, and number of matched points with SuperGlue[1] that is a state-of-the-art model for Feature Matching.  Our model achieved competitive results in terms of accuracy, and significantly outperformed SuperGlue in terms of number of correctly matched points (~50\%).

\section{Methods}\label{sec:method}
\begin{figure*}[!t]
    \vspace{-1.2cm}
    \centering
    \includegraphics[width=0.98\linewidth]{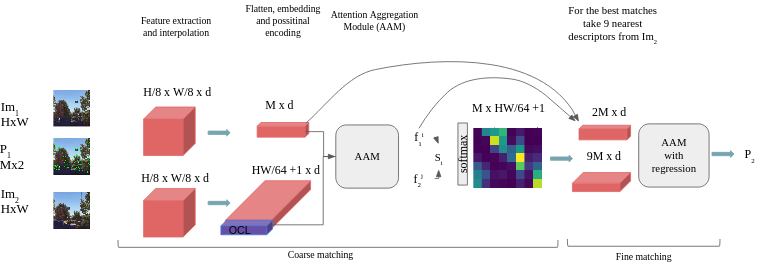}
    \caption{\textbf{Model architecture.}
        (Section \ref{sec:coarse}).
    }
    \label{fig:overview}
\end{figure*}

In this section, we present the details of the proposed approach for keypoints tracking using transformer networks.
Given two images $Im_1$, $Im_2$ of the size $H\times W$, and the positions of the points from the first image $P_1$ with the size $M\times2$ , we aim to find the location of corresponding points $P_2$ in the second image.  Instead of feature matching we propose to directly find the location of the corresponding keypoints in the $Im_2$. This allows us to skip the search of keypoints (at least in the second image), and thus our method does not depend on repeatability of searched keypoints. We divide the search of corresponding points $P_2$ in two stages called coarse matching, and fine matching, (fig.2). 

PUT FIGURE

During the first stage we intend to find the approximate position of  $P_2$, and discard outliers, with occluded points.  The second stage’s goal has been designed to find the precise position of $P_2$.  During the coarse matching we firstly extract image features from $Im_2$, and keypoints descriptors from $Im_1$, after we apply embeddings, with positional encoding, and finally process those features with Attention Aggregation Module (AAM). For the fine matching we apply Attention Aggregation Module to keypoint’s descriptors and features after the coarse matching with image features, and regress the result in order to get the final position of $P_2$. Further we will describe each step in detail.

\subsection{Feature extraction}\label{sec:local}
For the image features extraction, we used the CNN model architecture from SuperPoint(SP) [2]. This model proved to be one of the best for representative feature extraction. For an image of size $H\times W$ it extracts $\nicefrac{1}{8}\times\nicefrac{1}{8}\times 256$ features. Each feature characterizes a $8\times8$ patch around it, so we set its position to be the center of that patch. After the features are flattened, we form a set of $\nicefrac{W\times H}{64}$ pairs, where each entry $(f_i,p_i)$ represents the $i^{th}$ patch’s descriptor, and position. The process is visualized at fig. 3.

PUT FIGURE

For points’ feature descriptors, we firstly use SuperPoint to extract features from $Im_1$, and next based on points' position we use bi-linear interpolation to get their descriptors. As a results we obtain sparse features descriptors from the $Im_1$ and a dense feature representation of the second image in its entirety.

\subsection{Feature transformer}\label{sec:local}
During the next step, previously extracted features are passed to a positional encoding module, as well as to the Attentional aggregation modules. 
To provide a better understanding of Attention Layers, we firstly introduce the concept of transformer networks, and  explain the intuition behind them. 
A Transformer encoder consists of sequentially connected encoder layers, the core element of the encoder layer is the Attention layer. Traditionally the names of inputs to it are : $Q$ - Queries, $K$ - Keys, $V$ - Values. Assuming that the size of all matrices is Nd, we can define attention as : 
\begin{equation*}
    \text{Attention}(Q, K, V) = \operatorname{softmax}(QK^T)V.
    \label{eq:attn}
\end{equation*}

\begin{figure}[tb]
    \centering
    \includegraphics[width=0.95\linewidth]{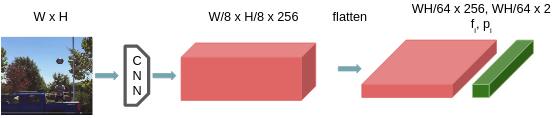}
    \caption[Features extraction.]{
        \textbf{Features extraction.}
    }
    \label{fig:teaser}
\end{figure}
PUT FIGURE
The intuition at this point is that attention operation learns feature level association by calculating similarity between Query and Key matrices. Another possible explanation is that by taking the $QK^T$ product we dynamically generate weights based on input of $Q$ and $K$ for linear layer with input $V$. In such a way that it is possible to increase the generalizability of the model as certain weights are generated on the fly and depend on the input. 

The main problem of attention layers in Computer Vision field is that their complexity grows quadratically $(O(N^2))$ as the length of the input increases (this can be clearly seen from fig. 4:

the size of the output from the first MatMul operation is basically $N^2$). In computer vision the input mainly consists of pixels,  patches, or their descriptors, so the complexity of the attention layer for images of size $(H,W)$ becomes proportional to $(HW)^2$. For example, even for $32\times32$ images this number becomes ~ $10^6$. Recently there were several successful attempts to challenge this problem [3] by expressing self-attention as a linear dot-product of kernel feature maps. In this project, we are taking advantage of this method to speed-up, and reduce the memory consumption of the model. The difference (see fig. 4) 

PUT FIGURE

\begin{figure}[tb]
    \centering
    \includegraphics[width=0.95\linewidth]{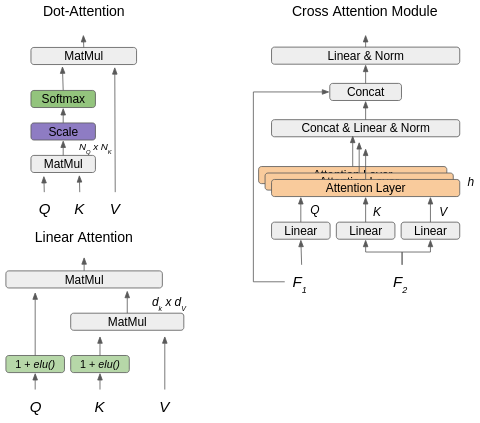}
    \caption[Comparison of Dot-attention with Linear Attention, and Structure of Cross Attention Module.]{
        \textbf{Comparison of Dot-attention with Linear Attention, and Structure of Cross Attention Module.}
    }
    \label{fig:teaser}
\end{figure}

is that in linear transformers instead of computing similarity between $Q$, and $K$ as $\operatorname{softmax}(QK^T)$,it is approximated as $\phi(Q) \cdot \phi(K)^T, \text{where } \phi(\cdot) = \operatorname{elu}(\cdot) + 1$.
Therefore, there is no need to compute a costly $\operatorname{softmax}$ operation, instead  $\phi(K)^T \cdot V$ is computed, which allows us to avoid large $N\times N$ matrices, while preserving the order of matrix multiplication.

\subsection{Positional encoding }\label{sec:local}
Positional encoding is of utter importance for any transformer-based model. In this work, we used Multi Layer Perceptron (MLP) to embed points' 2d positions into vectors with the dimension same to descriptors’ dimension, and then add them together, as you can see at fig. 5. 

PUT FIGURE 

\begin{equation*}
    f_i =f_i +  \operatorname{MLP}(p_i)
    \label{eq:pose_enc}
\end{equation*}

After features from the second set are concatenated with a special OCL token,  it can be treated as an additional feature from the $Im_2$. OCL token stands for occlusion token, it is used for finding occluded points, and outliers. This will be discussed in detail in further sections. 

\subsection{Attention aggregation module }\label{sec:local}

The attention aggregation module consists of several ($N_L$) self-attention, and cross-attention layers stacked together (see fig. 6).

The structure of the cross attention layer is shown on the fig (see fig. 4) .

The  input is two sets of features after positional encoding $F_1$ and $F_2$ or $F_2$ and $F_1$ depending on the direction of cross-attention. Self-attention layer has the exact same structure, the only difference is that we use 2 identical sets of features: $F_1$ and $F_1$ or $F_2$ and $F_2$  ( note that we consider OCL token as a part of $F_2$).  The primary goal of the self-attention layer is to share the information inside the same set of features, and the goal of the cross-attention layer is to share information between $F_1$ and $F_2$.

\begin{figure}[tb]
    \centering
    \includegraphics[width=0.65\linewidth]{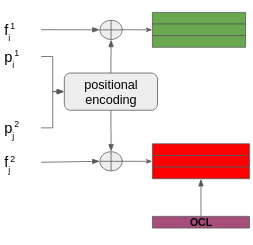}
    \caption[ Architecture of positional encoding module     ]{
        \textbf{  Architecture of positional encoding module     }
    }
    \label{fig:teaser}
\end{figure}
\subsection{Coarse matching}\label{sec:coarse}
In order to establish the coarse matching between $F_1$ and $F_2$, we calculate the similarity score $s_{ij}$ for each pair of features $f_1^i$ and $f_2^j$ by computing their dot product. In such a way we build a similarity matrix $S$ of the size $M \times\frac{HW}{64}+1$, and next apply softmax along the X direction (along direction with $\frac{HW}{64}+1$ points ) to obtain the probabilities of matching for each of $M$ keypoints.

\begin{equation*}
    S_{ij} = <f_1^i, f_2^j>,  \forall (i,j) \in M \times\frac{HW}{64}+1
    \label{eq:pose_enc}
\end{equation*}

Now the problem became similar to a multi label classification problem, we aim to classify each point from $F_1$ to one of $\frac{HW}{64}+1$ classes. 
We select matches based on the highest similarity score. Optionally, it is possible to consider the match if the similarity score is higher than some threshold. Point is  discarded if its confidence score is lower than threshold or if it's best match is OCL token.

\subsection{Fine matching }\label{sec:local}
After establishing the coarse matching, we know the approximate position of the points on $Im_2$, and we assume that the exact match is located in the $8\times8$ region around the matched feature. 
Since we are using bi-linear interpolation to extract the point descriptors, we can write $f_1^j$ as linear combination of $F_1^i$, where $F_1^i$ are the nearest descriptors to the point on the $Im_1$ (see fig. 7).

Combination of $\alpha_i$ - coefficients of bi-lienar interpolation, uniquely determines the relative position of the $f_1^j$ :

\begin{equation*}
    f_1^j = \sum_{i \in (1,4)} \alpha_iF_i^i, \quad s.t. \sum{}\alpha_i = 1
    \label{eq:pose_enc}
\end{equation*}

\begin{figure}[tb]
    \centering
    \includegraphics[width=0.95\linewidth]{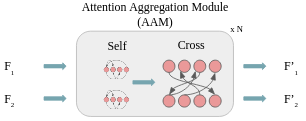}
    \caption[  Structure of Attention Aggregation Module  .]{
        \textbf{ Structure of Attention Aggregation Module  .}
    }
    \label{fig:teaser}
\end{figure}

Thus we can assume that it is also possible to express $ f_1^j$ as a linear combination of the nearest descriptors from $Im_2$  with some noise $n_j$:
\begin{equation*}
    f_1^j = \sum_{i \in (1,9)} \alpha_iF_i^i + n_j
    \label{eq:pose_enc}
\end{equation*}
Here we use 9 $F_2^i$ (see fig. 7)

FIG REFERENCE

as we do not know the exact location of the point on the second image. Again, we can assume that coefficients $\alpha_i$ determine the points’ positions. Thus based on these assumptions, for each keypoint we use its original descriptor with representation after AAM, and take 9 nearest image descriptors from Im2. Next we process them using the Attention aggregation module, calculate the similarity scores, and use them to regress the relative position of the point (see fig. 8 for details)

that allows us to move to sub-pixel matching. 

\subsection{OCL Token}\label{sec:local}

\begin{figure}[tb]
    \centering
    \includegraphics[width=0.75\linewidth]{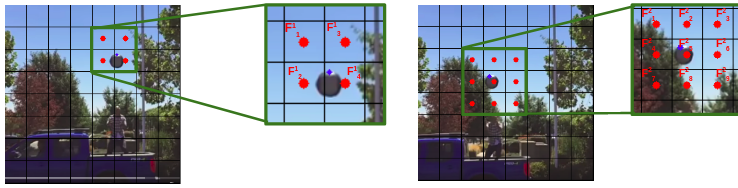}
    \caption[ Illustration of principles for fine module  ]{
        \textbf{ Illustration of principles for fine module  }
    }
    \label{fig:teaser}
\end{figure}

One of the biggest challenges of points and images matching is to deal with outliers, and occluded points. The popular approach to deal with this problem, in the context of deep learning based keypoints matching, is the Sinkhorn Algorithm[21]. After computing the similarity matrix, this approach formulates the matching problem as a bipartite assignment solving the optimal transport problem. 
However, it can not be used for point tracking, as there is a chance that two different points from $F_1$ are matched to the same patch from $F_2$ (multiple to one matching case). Therefore, we employ another strategy to discard occlusions: we concatenate the $F_2$ set with a learnable OCL token. Simply speaking, we just add one additional feature to $F_2$ that stands for patch outside the $Im_2$ . During the training stage, the model is taught to match all occluded points to OCL token. Intuitively, the coarse module processes the features of the OCL token in a way that it becomes quite similar to all features in $F_1$ and different from all features from F2. Thus, after computing the similarity matrix, all features $f_1$ that lack similar features $f_2$ are classified as outliers. 
As we discussed in the Coarse matching part, after calculating the similarity matrix, the problem can be formulated as a classification. 
Let assume that 20\% of the keypoints are occluded, then the probability that one point is matched with an OCL token is 0.2, while its probability to be matched with any other feature is:
\begin{equation*}
    \text{Prob}(p_i \text{ is not occluded}) = \frac{0.8* M*64}{HW}, M<<WH
    \label{eq:pose_enc}
\end{equation*}

This means that our dataset is extremely unbalanced, and direct training will lead to classifying all points as occluded. Thus we divided the training into several stages, this will be discussed in detail in the Training strategy section.

\begin{figure}[tb]
    \centering
    \includegraphics[width=0.95\linewidth]{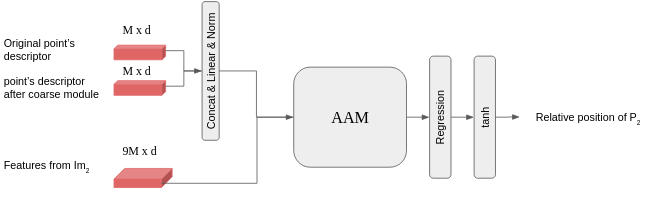}
    \caption[ Architecture of Fine matching module  ]{
        \textbf{Architecture of Fine matching module }
    }
    \label{fig:teaser}
\end{figure}

\section{Dataset Generation }\label{sec:data}

Since there is no large dataset of images with pixel to pixel correspondence for sparse matching, we created a synthetic dataset of geometric primitives, and a real dataset of warped images. The synthetic dataset is used to warm up the training, as real images appear to be too complex to ensure the convergence of the network without a proper initialization.

\subsection{Synthetic dataset}\label{sec:local}

The Synthetic dataset consists of image pairs.  Specifically, each image consists of a background full of stripes, triangles, stars, etc., and a big cube located in the center of the image. The first, and second images differ by small displacement of background(~16 pixels), and  huge displacement of the cube(~ 50 pixels ).

For this dataset, we select the corners of the geometric primitives as points to track. While the structure of the images is rather simple, it is still a considerably difficult task, as there is a need to track points with different levels of displacement simultaneously. Thus, such an approach disables the model to learn homographies instead of point tracking. 
Since the displacement of the cube is large compared to the displacement of the background, it is likely that some points on the second image naturally become occluded. Hence, we use those to train OCL token. 
 
\subsection{Real dataset}\label{sec:local}
For training on the real images we select the COCO14[23] image dataset. For each image we 1) applied random projective transformation to the image, 2) selected regions with huge concentration of key points, 3) cropped patches of the same size that are centered at corresponding points, 4) selected key points on the first patch, and used them as P1; those of them that do not appear on the second patch are treated as outliers.

\section{Training strategy}\label{sec:data}
\begin{figure}[tb]
    \centering
    \includegraphics[width=0.95\linewidth]{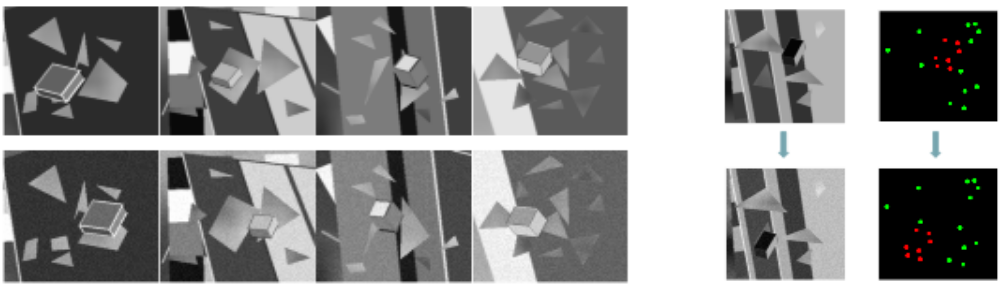}
    \caption[ Examples of Synthetic dataset. The right figure demonstrates the displacement of the Cube   ]{
        \textbf{  Examples of Synthetic dataset. The right figure demonstrates the displacement of the Cube  }
    }
    \label{fig:teaser}
\end{figure}
 We divide the training into 3 stages: 
 
\subsection{Synthetic dataset without occlusions}\label{sec:local}

Firstly we trained the model on the Synthetic dataset without occluded points in order to “warm up” the model. For each keypoint $p_1^i$, the ground truth consists of the index of the patch where this keypoint is located  - $y^i$, and its position on the second image - $p_2^i$. Based on the predicted similarity score $s^i$, it is possible to select predicted position on the second patch - $\tilde{p_2}^i$.

\begin{equation*}
    s_{i} = <f_1^i, f_2^j>,  \forall  f_2^j \in F
    \label{eq:pose_enc}
\end{equation*}

Thus the total loss consists of Cross Entropy Loss between $s^i$ and $y^i$, also, we added the L2 loss between $p_2^i$ and  $p_2^i$ in order to speed up the convergence time:

\begin{equation*}
    \mathcal{L}_{total} = \sum_{i \in (1,M)}CEL(s^i,y^i)  + L2(p_2^i,\tilde{p_2}^i)
    \label{eq:pose_enc}
\end{equation*}

\subsection{Synthetic dataset with occlusions}\label{sec:local}
At this stage we used the same dataset, however now occluded points are also used, and the loss consists only of CEL:

\begin{equation*}
    \mathcal{L}_{total} = \sum_{i \in (1,M)}CEL(s^i,y^i) 
    \label{eq:pose_enc}
\end{equation*}

This transition stage is extremely important, as ‘occluded class’ should be added only after the model is taught how to distinguish between all inliers due to reasons we discussed in the Coarse matching part.

\subsection{Training on a dataset with real images}\label{sec:local}

Finally, we train the model on the dataset of real images. At this stage we applied random color jitter to the images to mimic the change of illumination, and used projective transforms of different difficulty levels. The loss again consists only of CEL:
\begin{equation*}
    \mathcal{L}_{total} = \sum_{i \in (1,M)}CEL(s^i,y^i) 
    \label{eq:pose_enc}
\end{equation*}

\subsection{Training fine matching module}\label{sec:local}

For training the fine matching module, we freeze the weights of the Coarse module. Based on the results of coarse module, for each point $p_1^i$ and position of the patch where it is located ($p_2^i$), we predict their relative distance $d^* \in (-4,4)$, having the ground truth $d$.  For this we use L2 loss:
\begin{equation*}
    \mathcal{L}_{total} = \sum_{i \in (1,M)}L2(d,d^*) 
    \label{eq:pose_enc}
\end{equation*}
Thus the result position of the keypoint becomes: $\tilde{p_2}^i+d^*$  .

\begin{figure}[tb]
    \centering
    \includegraphics[width=0.95\linewidth]{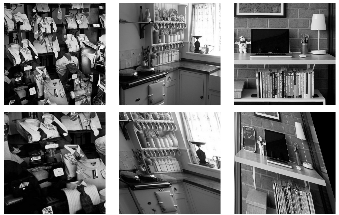}
    \caption[  Examples of real dataset  ]{
        \textbf{ Examples of real dataset   }
    }
    \label{fig:teaser}
\end{figure}

\section{Results}\label{sec:data}

Since there are no analogous works, we provide a comparison with the state-of-the-art method for keypoint feature matching SuperGlue[1]. We tested our model on 2 datasets: COCO2014, and HPatches under different levels of  light intensity, and projective transforms. For the evaluation metrics we selected 1) matching accuracy: the match is correct if distance between predicted and real position of the point is less than 6 pixels, and 2) number of correctly matched keypoints. 

\subsection{Evaluation Protocol}\label{sec:local}

In order to provide a fair comparison, we conducted the experiments in the following way: for each image in the dataset 1) we created a corresponding image by applying projective transforms, and color jitter 2) sampled 512 most representative keypoints from the first image - $kp_1$, and  all keypoints from the second image - $kp_2$ 3) matched $kp_1$ to $kp_2$ using SuperGlue, and matched $kp_1$ to the entire $Im_2$ 4) compare the results. The table below summarizes the outcomes of our results (Table 1, Table 2).

\begin{table}[tb]
    \centering
    \scriptsize{
        
\setlength\tabcolsep{4.0pt}
\begin{tabular}{clcccc}
    \toprule
    \multirow{2}{*}[+.4em]{Dataset} & \multicolumn{1}{c}{\multirow{2}{*}[+.4em]{SuperGlue}} & \multicolumn{1}{c}{\multirow{2}{*}[+.4em]{Our model(coarse only)}}  &
    \multicolumn{1}{c}{\multirow{2}{*}[+.4em]{Our model}}
    \\

    \midrule
    \multirow{1}{1.5cm}[-.0em]{COCO easy} & 94.8   & 93.5     & \b{95.3}            \\
    \multirow{1}{1.5cm}[-.0em]{COCO hard} & 91.6   & 90.2     & \b{91.7}            \\
    \multirow{1}{1.5cm}[-.0em]{COCO illum} &\b{89.0}   & 87.2     & 88.8            \\

    \midrule
    \multirow{1}{1.5cm}[-.0em]{Hpatches hard} 91.4   & 90.5     & \b{91.5}       \\
    \bottomrule
\end{tabular}
    }
    \caption{\b{Accuracy results}}
    \label{tab:megadepth}
\end{table}

\begin{table}[tb]
    \centering
    \scriptsize{
        
\setlength\tabcolsep{4.0pt}
\begin{tabular}{clcccc}
    \toprule
    \multirow{2}{*}[+.4em]{Dataset} & \multicolumn{1}{c}{\multirow{2}{*}[+.4em]{SuperGlue}} & \multicolumn{1}{c}{\multirow{2}{*}[+.4em]{Our model}}  
    \\

    \midrule
    \multirow{1}{1.5cm}[-.0em]{COCO easy} & 249   & 358             \\
    \multirow{1}{1.5cm}[-.0em]{COCO hard} & 240  &  346           \\
    \multirow{1}{1.5cm}[-.0em]{COCO illum}& 195   & 300             \\

    \midrule
    \multirow{1}{1.5cm}[-.0em]{Hpatches hard} & 222   & 340         \\
    \bottomrule
\end{tabular}
    }
    \caption{\b{Number of correctly  matched  point out of 512}}
    \label{tab:megadepth}
\end{table}
\begin{figure}[tb]
    \centering
    \includegraphics[width=0.95\linewidth]{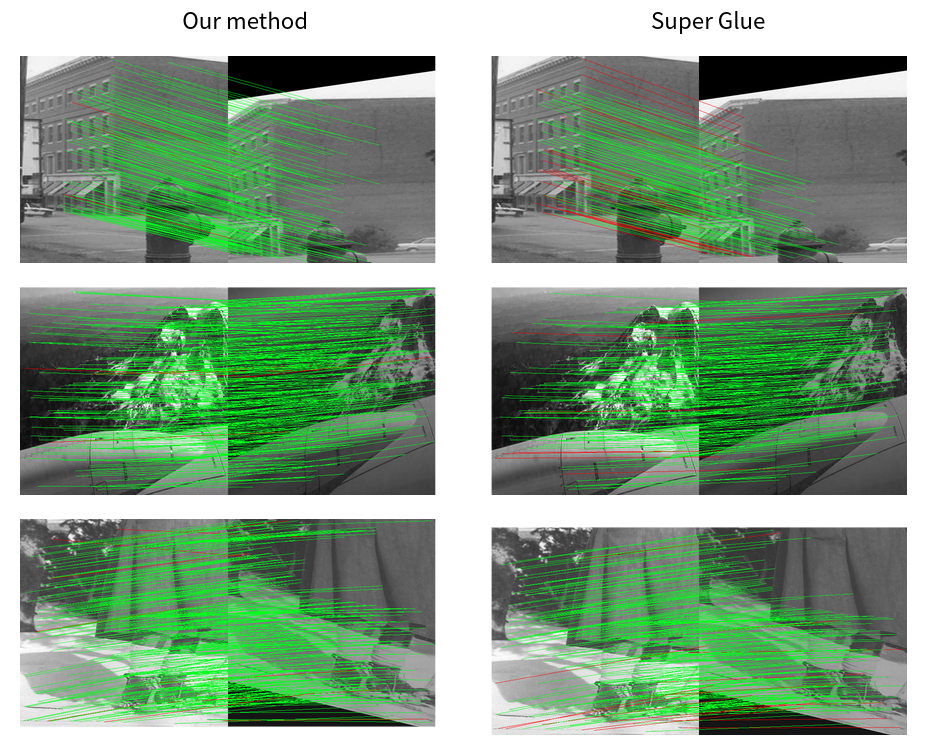}
    \caption[ Qualitative results. Our model is compared to SuperGlue. Our method demonstrates a higher number of matched keypoints.  e   ]{
        \textbf{ Qualitative results. Our model is compared to SuperGlue. Our method demonstrates a higher number of matched keypoints.   }
    }
    \label{fig:teaser}
\end{figure}
During testing of our model without a fine matching module, we considered the predicted patch’s center as the prediction for keypoints’ positions. Since we put the distance threshold to be 6 pixels, the accuracy of the coarse module represents the accuracy of the matching point to the correct patch. The fine module on average adds ~1.5\% accuracy, the reason is that for the cases when the coarse module predicts a nearby patch instead of the correct one, the fine module may refine the predicted position in a way that it becomes closer to the correct patch.
The results clearly show that our model reaches the same, and sometimes higher level of accuracy compared to SuperGlue, while producing a significantly larger number of correctly matched points

\subsection{Limitations and future work directions}\label{sec:local}
Given two images and input points from one of them, our current method finds the position of these points on the second image. To further improve the matching accuracy, we can additionally try to find location of the predicted points on the first image, in such a way it is possible to increase robustness, and establish semi-supervised learning technique. Another way to improve the current results is to experiment with different numbers of attention layers, and train models on image sequences from Megadepth or ScanNet Datasets. In such a way the models can learn different sets of parameters for indoor, and outdoor environments. Additionally, instead of separate training of coarse and fine modules, we could integrate it in a single training pipeline that can be used  out of the box in SLAM. We plan to extend this work to journal or conference publication. 

\section{Conclusion}\label{sec:data}
We have proposed a novel deep learning architecture which ensures a robust keypoints tracking implemented in an hierarchical manner. It consists of two successive stages, a coarse matching followed by a fine localization of the keypoints’ correspondances prediction. Through various experiments, we demonstrated that our approach achieves competitive results and demonstrates high robustness against adverse conditions, such as illumination change, occlusion and viewpoint differences while producing a significantly larger number of correctly matched keypoints, compared to keypoints matching techniques.  
Moreover, as a pioneering work introducing transformer networks for keypoints tracking tasks, we believe that our research can be the beginning of a novel research track.
\pagebreak[2]

\section{Reference Literature}\label{sec:data}

1 Paul-Edouard Sarlin, Daniel DeTone, Tomasz Malisiewicz,and  Andrew  Rabinovich.SuperGlue:   Learning  feature matching with graph neural networks.   CVPR, 2020 \\
2 Daniel  DeTone,  Tomasz  Malisiewicz,  and  Andrew  Rabinovich.  SuperPoint: Self-supervised interest point detection and description. CVPRW, 2018 \\
3 Angelos Katharopoulos, Apoorv Vyas, Nikolaos Pappas, and Francois Fleuret.  Transformers are RNNs:  Fast autoregressive transformers with linear attention.  ICML, 2020 \\
4 Daniel  DeTone,  Tomasz  Malisiewicz,  and  Andrew  Rabinovich.   Toward geometric deep slam.arXiv:1707.07410. \\
5 Zichao Zhang, Torsten Sattler, and Davide Scaramuzza. Reference Pose Generation for Long-term Visual Localization Via Learned Features and View Synthesis. IJCV, 2020
6  Raul Mur-Artal, Juan D. Tardos. ORB-SLAM2: an Open-Source SLAM System for Monocular, Stereo and RGB-D Cameras. IEEE 2017 \\
7 Junjie Huang, Wei Zou, Zheng Zhu, Jiagang Zhu. An Efficient Optical Flow Based Motion Detection Method for Non-stationary Scenes.    arXiv:1811.08290 \\
8 Aashi Manglik, Xinshuo Weng, Eshed Ohn-Bar, Kris M. Kitani. Forecasting Time-to-Collision from Monocular Video: Feasibility, Dataset, and Challenges. IROS, 2019. \\
9 Sourav Garg, Ben Harwood, Gaurangi Anand, Michael Milford. Delta Descriptors: Change-Based Place Representation for Robust Visual Localization. IEEE Robotics and Automation Letters (RA-L), 2020 \\
10 David  G  Lowe.Distinctive  image  features  from  scale-invariant keypoints. IJCV, 2004. \\
11 Ethan  Rublee,  Vincent  Rabaud,  Kurt  Konolige,  and  GaryBradski.   ORB:  An  efficient  alternative  to  SIFT  or  SURF. In ICCV, 2011 \\
12 Jerome Revaud, Philippe Weinzaepfel, Cesar De Souza, Noe Pion, Gabriela Csurka, Yohann Cabon, and Martin Humenberger. R2D2: repeatable and reliable detector and descrip-tor. NeurIPS, 2019 \\
13 Ignacio Rocco, Relja Arandjelovicc, and Josef Sivic. Efficient Neighbourhood consensus networks via submanifold sparse convolutions. In ECCV, 2020 \\
14 Chhaniyara, Savan; KASPAR ALTHOEFER; LAKMAL D. SENEVIRATNE . Visual Odometry Technique Using Circular Marker Identification For Motion Parameter Estimation. Advances in Mobile Robotics, 2008 \\
15  Igor Vasiljevic, Vitor Guizilini, Rares Ambrus, Sudeep Pillai, Wolfram Burgard, Greg Shakhnarovich, Adrien Gaidon. Neural Ray Surfaces for Self-Supervised Learning of Depth and Ego-motion. International Conference on 3D Vision, 2020 \\
16 Kwang Moo Yi, Eduard Trulls, Vincent Lepetit, Pascal Fua. LIFT: Learned Invariant Feature Transform. ECCV 2016 \\
17 Mihai Dusmanu, Ignacio Rocco, Tomas Pajdla, Marc Pollefeys, Josef Sivic, Akihiko Torii, Torsten Sattler \\
18  Zhengqi Li and Noah Snavely. Megadepth: Learning single-view depth prediction from internet photos.  CVPR, 2018 \\
19 Angela Dai,  Angel X Chang,  Manolis Savva,  Maciej Hal-ber, Thomas Funkhouser, and Matthias Niessner.   ScanNet:Richly-annotated  3d  reconstructions  of  indoor  scenes. CVPR, 2017 \\
20 Ashish Vaswani, Noam Shazeer, Niki Parmar, Jakob Uszkoreit, Llion Jones, Aidan N Gomez, Lukasz Kaiser, and IlliaPolosukhin. Attention is all you need. NeurIPS, 2017 \\
21 Marco Cuturi.  Sinkhorn distances: Lightspeed computation of optimal transport. NIPS, 2013 \\
22 Jay S., Rabatel G., Hadoux X., Moura D. and Gorretta N. In-ﬁeld crop row phenotyping from 3D modeling performed using Structure from Motion. Computers and Electronics in Agriculture, 110, 70-77, 2015 \\
23  T.-Y. Lin, M. Maire, S. Belongie, J. Hays, P. Perona,D.  Ramanan,  P.  Dollar,  and  L.  Zitnick.    Microsoft. COCO: Common objects in context.  ECCV, 2014

{\small
\bibliographystyle{ieeefullname}
\bibliography{fulldb_cleaned_noarxiv}
}

\end{document}